\documentclass[letterpaper, 10 pt, conference]{ieeeconf}  

\IEEEoverridecommandlockouts                              

\usepackage{subcaption}

\usepackage{cite}
\usepackage{amsmath,amssymb,amsfonts}
\usepackage{hyperref} 
\usepackage{algorithm}
\usepackage{algpseudocode}
\usepackage{graphicx}
\usepackage{textcomp}
\usepackage{xcolor}
\def\BibTeX{{\rm B\kern-.05em{\sc i\kern-.025em b}\kern-.08em
    T\kern-.1667em\lower.7ex\hbox{E}\kern-.125emX}}
\begin{document}

\captionsetup[figure]{font=footnotesize}

\title{\LARGE \bf
BeadSight: An Inexpensive Tactile Sensor Using Hydro-Gel Beads
}

\author{Abraham George$^{1}$, Yibo Chen$^{1}$, Atharva Dikshit$^{1}$, Peter Pak$^{1}$, and Amir Barati Farimani$^{1}$ 
\thanks{$^{1}$With the Department of Mechanical Engineering,
        Carnegie Mellon University 
        {\tt\small \{aigeorge, yibochen, adikshit, ppak, afariman\} @andrew.cmu.edu}}%
}


\maketitle
\thispagestyle{empty}
\pagestyle{empty}

\begin{abstract}

In robotic manipulation, tactile sensors are indispensable, especially when dealing with soft objects, objects of varying dimensions, or those out of the robot's direct line of sight. Traditional tactile sensors often grapple with challenges related to cost and durability. To address these issues, our study introduces a novel approach to visuo-tactile sensing with an emphasis on economy and replacablity. Our proposed sensor, BeadSight, uses hydro-gel beads encased in a vinyl bag as an economical, easily replaceable sensing medium. When the sensor makes contact with a surface, the deformation of the hydrogel beads is observed using a rear camera. This observation is then passed through a U-net Neural Network to predict the forces acting on the surface of the bead bag, in the form of a pressure map. Our results show that the sensor can accurately predict these pressure maps, detecting the location and magnitude of forces applied to the surface. These abilities make BeadSight an effective, inexpensive, and easily replaceable tactile sensor, ideal for many robotics applications.


\end{abstract}


\section{Introduction}


In the ever-evolving realm of robotics, the ability to interact with a diverse array of objects has become paramount. The foundation of this capability lies in the robot's sense of touch. Tactile sensors \cite{YOUSEF2011171}, therefore, emerge as indispensable tools in advancing the frontiers of robotic manipulation, especially in environments that are intricate, unpredictable, or both  \cite{ma2019dense} \cite{8202149}. 


Recently, large strides in tactile sensing have been made using visuo-tactile sensors. This family of sensors, which was pioneered by GelSight, measures contact forces by visually tracking the deformation of a contact surface \cite{yuan2017gelsight} \cite{taylor2022gelslim}. These sensors feature a deformable membrane, often crafted from silicon, which serves as the working surface of the sensor. When this surface makes contact, a rear camera measures its deformation using tracking markers and directional lighting. The applied forces can be calculated from these measured deformations based on known material properties of the sensor pad.

\begin{figure}[thpb]
    \centering
    \includegraphics[width=\columnwidth]{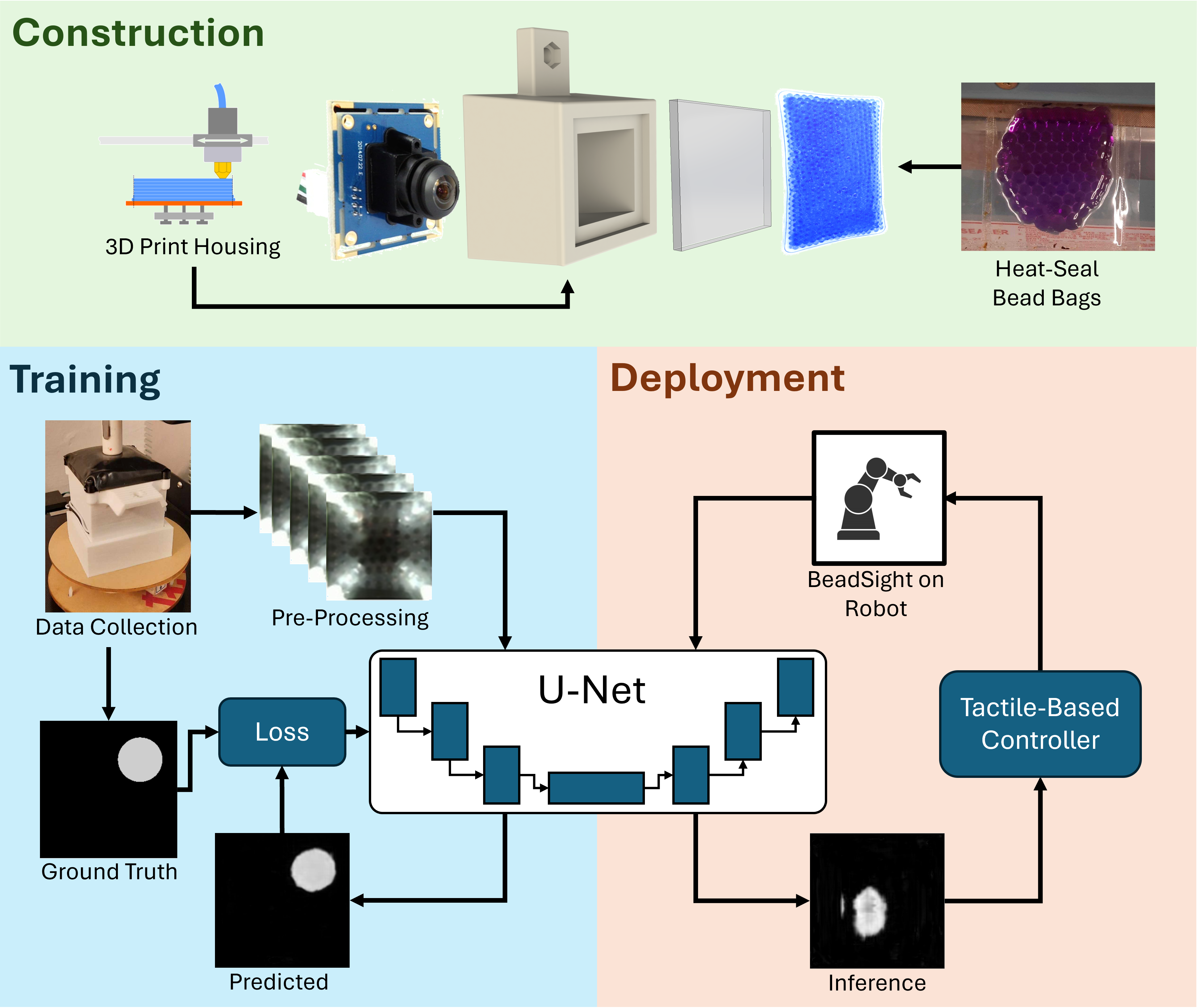}    \caption{\label{fig:overview}BeadSight Overview. The sensor is composed of a camera mounted inside of a 3D printed housing, observing the deformations of hydro-gel beads encased in a vinyl pouch. A U-Net was trained to recreate contact pressure maps from the camera observations. During deployment, the bead sight can be mounted to a robot, and the recreated pressure maps can be used for applications, such as control.}
    \label{fig:overview}
\end{figure}

Although these visuotactile sensors can determine the deformation of the surface, and therefore the applied forces, with great accuracy, they face a few critical limitations. Primarily, their design leads to issues of durability. Over time and with sustained use, the sensor pad may become damaged, leading to drifts in readings.  Leading visual-tactile sensors, such as gelsight, prioritize high-quality reconstructions over durability and therefore use soft, highly deformable silicon pads as a sensor surface. The lack of emphasis on durability can be especially problematic in robotic settings, where contact forces can be large and control algorithms can be aggressive. The reliability of the sensor, especially in rugged or prolonged applications, remains a concern. For example \cite{george2024visuo} trained an imitation learning agent to plug in a cable, using tactile information from a GelSight sensor. However, the high contact forces of the plugging task caused the senor pad to rip after only 20 to 30 insertion attempts, causing the researchers to create a silicon cover for the GelSight to increase durability. With that cover, they were able to achieve around 200 to 300 insertions before failure.  Compounded by the issue of durability, cost is another major limiting factor. Current tactile sensors are moderately expensive, with a GelSight mini costing \$499 for the sensor and \$69 for each replacement pad \cite{gelsightstore}. However, due to durability concerns, the cost of replacement pads for the sensor can become a limiting factor for many applications. Looking at the cable plugging example from \cite{george2024visuo}, without the additional cover, the replacement pads cost over \$2 per plugging attempt, making this technology impractical for many industrial applications.

To address these issues, we present a novel tactile sensor design that focuses on durability and affordability, while still managing to achieve a reasonable pressure map reconstruction. When designing our sensor, we specifically focused on the application of sensing for robotic control, with specific attention to the following areas:
\begin{enumerate}
    \item Grasp Confirmation and Localization: Grasping is a cornerstone of robotic manipulation and is a task in which tactile sensing can greatly help by providing essential feedback that confirms the success and location of a grasp \cite{wettels2009multi}. Although this task is simple, it is ubiquitous in robotic manipulation, and as such we focused on it during the development of our durable, inexpensive sensor. 
    \item Force Feedback: A delicate balance of force is crucial in robotic manipulation. Too much force can damage an object, while too little might result in a failed grasp. Tactile sensors offer the robot a nuanced understanding, enabling it to modulate its force appropriately \cite{9832483}. As such, our sensor design must be able to determine the magnitude and location of forces applied to it.
    \item Exploration: Robots often encounter unknown terrains and objects. Tactile sensors act as the robot's fingertips, helping it identify objects it comes in contact with \cite{she2021cable}. This tactile knowledge subsequently aids the robot in planning its actions, ensuring both safety and efficiency. Our design attempts to tackle this application area by providing detailed maps of the sensor's contact area.
\end{enumerate}

With these applications in mind, we present BeadSight, a tactile sensor that uses visual tracking of the deformation of a pouch of hydrogel beads to determine tactile interactions. The main benefit of this design is that the sensing surface, which is made of hydrogel beads incased in a 40mm by 40mm vinyl cover, is both durable and inexpensive. When contact forces are applied to the surface of the bead bag, the beads deform and move with-in the bag. These motions are captured by a rear camera and used to calculate the applied pressure map. Unfortunately, the beads' motion and deformation are stochastic and difficult to model, making the calculation of a pressure map from first principles unpracticable. Instead, we present a deep neural network architecture that is trained to reconstruct pressure maps from camera observations. Our results show that the BeadSight sensor, combined with a trained U\-Net model, can reconstruct the pressure maps of contacts with a mean absolute error of only 0.79 kPa. An overview of our proposed system can be seen in Fig. \ref{fig:overview}.

\section{Related work}\label{sec:related work}

\subsection{Tactile Sensors}
A wide range of tactile sensors have been developed in recent years for a large variety of robotic tasks \cite{li2020review}, ranging from object localization \cite{poseEstimation, 6697009}, to exploration \cite{tactileSearch, 7961193, 7837664}, to action control, both low-level feedback control \cite{DextrusInhandManipulation, 880799} and higher level planning \cite{6943031, 5980049}. These sensors approach the task of tactile sensing by measuring a variety of physical phenomena, all attempting to provide some understanding of the object the sensor is in contact with. The most common sensing modality is normal force detection, \cite{schurmann2011modular, zhang2010sensitivity} although this modality is often combined with the ability to measure tangential forces \cite{liu2015finger, tomo2017covering, ward2018tactip} or contact torques \cite{de2012force, palli2014dexmart}. Beyond these contact-force-based methods, tactile sensors have measured vibration \cite{meier2016distinguishing, zoller2018acoustic}, thermal contact \cite{siegel1986integrated, wade2015handheld}, and even object proximity \cite{schlegl2013pretouch, guo2015transmissive}. However, our sensor is mainly focused on normal force detection, as it is especially useful for grasping \cite{miller1999integration}. 

Visuo-tactile sensors have recently shown exceptional performance for contact force and surface deformation detection \cite{yuan2017gelsight}. This class of sensors, pioneered by GelSight \cite{yuan2017gelsight}, uses a camera, combined with tracking markers and directional lighting, to measure the deformation of a contact surface. Because the surface is illuminated with different colored lights at different orientations around the sensor surface, the observed color of the surface can be used to determine the normal direction of the sensor surface. The normal direction from the surface color is then integrated to reconstruct the surface and is combined with the motion of tracking markers on the surface to determine the sensor's total deformation. This deformation, when combined with known material properties of the sensing surface, can determine the exact forces applied to the surface \cite{8202149}. This method was adapted and improved upon to improve reconstruction accuracy  \cite{do2022densetact, do2023densetact} and reduce size \cite{do2023densetactmini}. Other approaches have added additional functionality to this type of visuotactile sensor, such as the See Through Your Skin sensor \cite{hogan2021seeing}, which used a semi-transparent silicon layer with tracking markers and an adjustable backlight to multi-task as both a tactile sensor (when the backlight is on) and a palm camera (when the backlight is off).

\subsection{U-Net Model for Visual Processing}

The U-Net is a type of convolutional neural network that was originally developed for biomedical image segmentation but has since been applied to a variety of image-to-image and image-to-mask tasks \cite{ronneberger2015u} \cite{qin2020u2} \cite{oktay2018attention}. The network is composed of two paths, an encoding path and a decoding path. The encoding path consists of repeated application of convolutional layers and max pooling operations \cite{10373700}, to iteratively condense a series of feature maps. This path is used to capture the high-level information present in the input. Meanwhile, the decoder path iteratively expands the feature map, eventually reaching the size of the goal output. Essentially, this encoder-decoder structure allows the network to extract high-level features, and then use them to recreate an output of a similar shape as the input. However, a pure encoder-decoder approach loses fine-grain features during the encoding process. To avoid this, \cite{37568589} proposed the use of skip connections, where outputs from earlier layers in the network are concatenated with the inputs to later layers, allowing the network to use information from multiple scales simultaneously.

\section{Tactile sensor design}\label{sec:sensor design}
\begin{figure}[htbp]
\centering
\includegraphics[width=0.8\columnwidth]{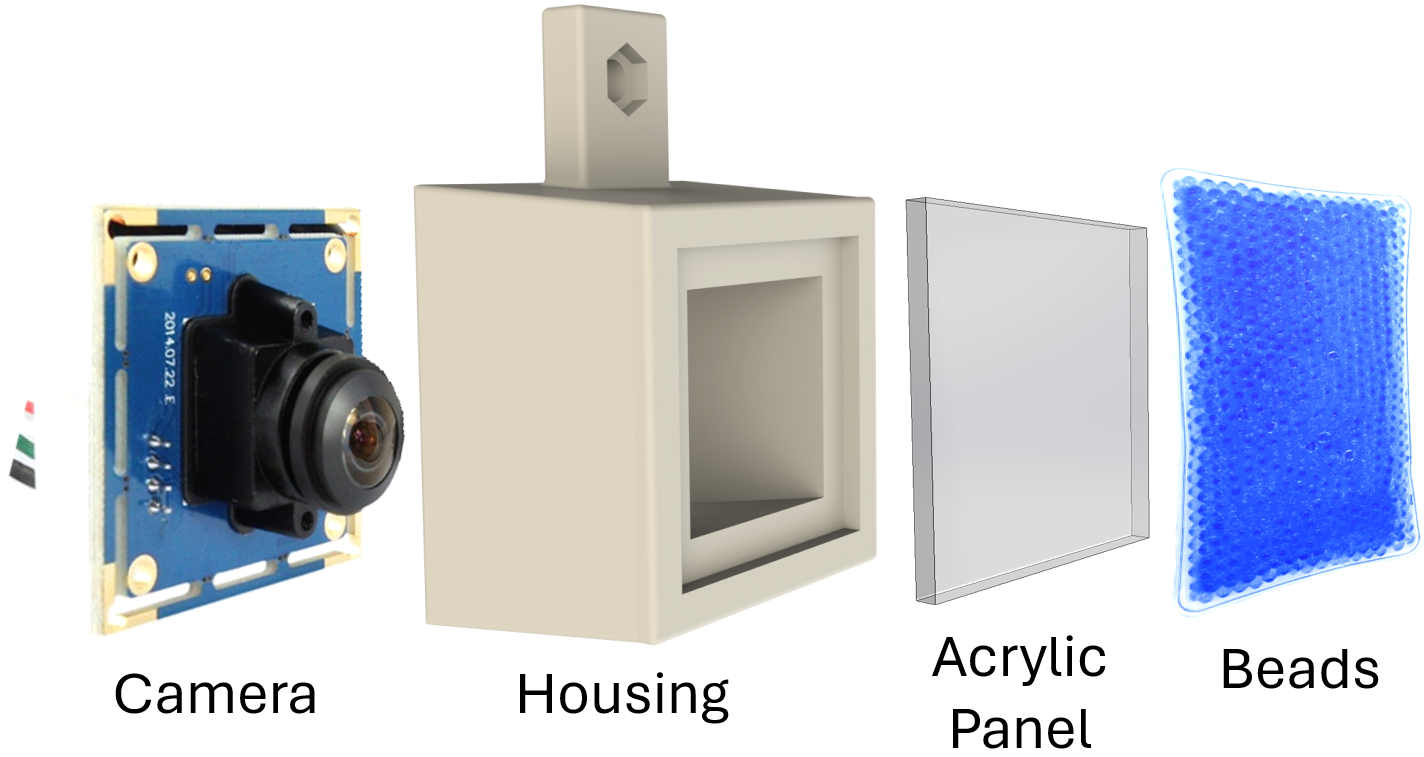}
\caption{BeadSight hardware. The camera is mounted at the rear of a 3D-printed housing. At the front, a 40mm x 40mm acrylic window supports the hydro-gel bead pouch.}
\label{fig:hardware_parts}
\end{figure}

The BeadSight sensor is built around a 3D printed housing, which holds a camera, four LEDs for illumination, and a 40 mm x 40 mm Polyvinyl Chloride (PVC) pouch resting against an acrylic pain (see Fig. \ref{fig:hardware_parts}). The pouch contains approximately 100 water beads, heat-sealed for an airtight bond. The 15 mil (0.381 mm) thick vinyl sheet provides a durable surface to withstand punctures while also enabling adequate flexibility to conform to sample surfaces. The water beads are composed of Polyacrylamide (PAM) with a 1.66 mm initial diameter. These beads were utilized for their known expansive and mechanical properties in various environments of osmotic pressure \cite{traber2019polyacrylamide}. Once saturated, these beads expand to around 4 mm in diameter. An additional 7 mil (0.178 mm) thick black PVC sheet was applied to the top of the bead sack for our experiments to visually isolate the BeadSight camera. However, this layer is optional and can be omitted if palm-camera observations would be helpful, such as if BeadSight is used for pixel-to-action machine-learning robotic control. 

\begin{figure}[htbp]
\centering
\begin{subfigure}{0.3\columnwidth}
    \centering
    \includegraphics[height=18mm]{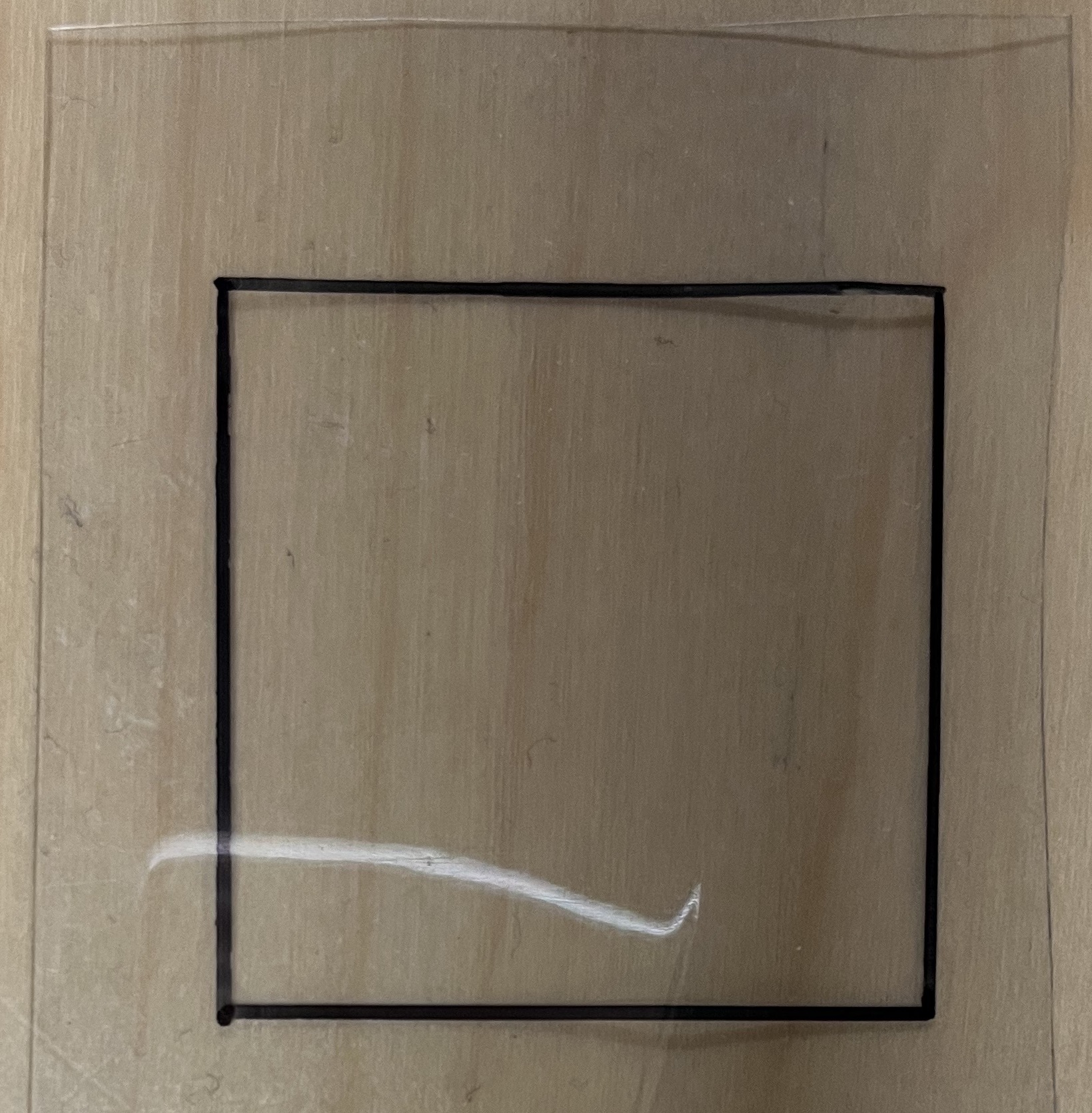}
    \caption{}
    \label{fig:bead_bag_outline}
\end{subfigure}
\begin{subfigure}{0.3\columnwidth}
    \centering
    \includegraphics[height=18mm]{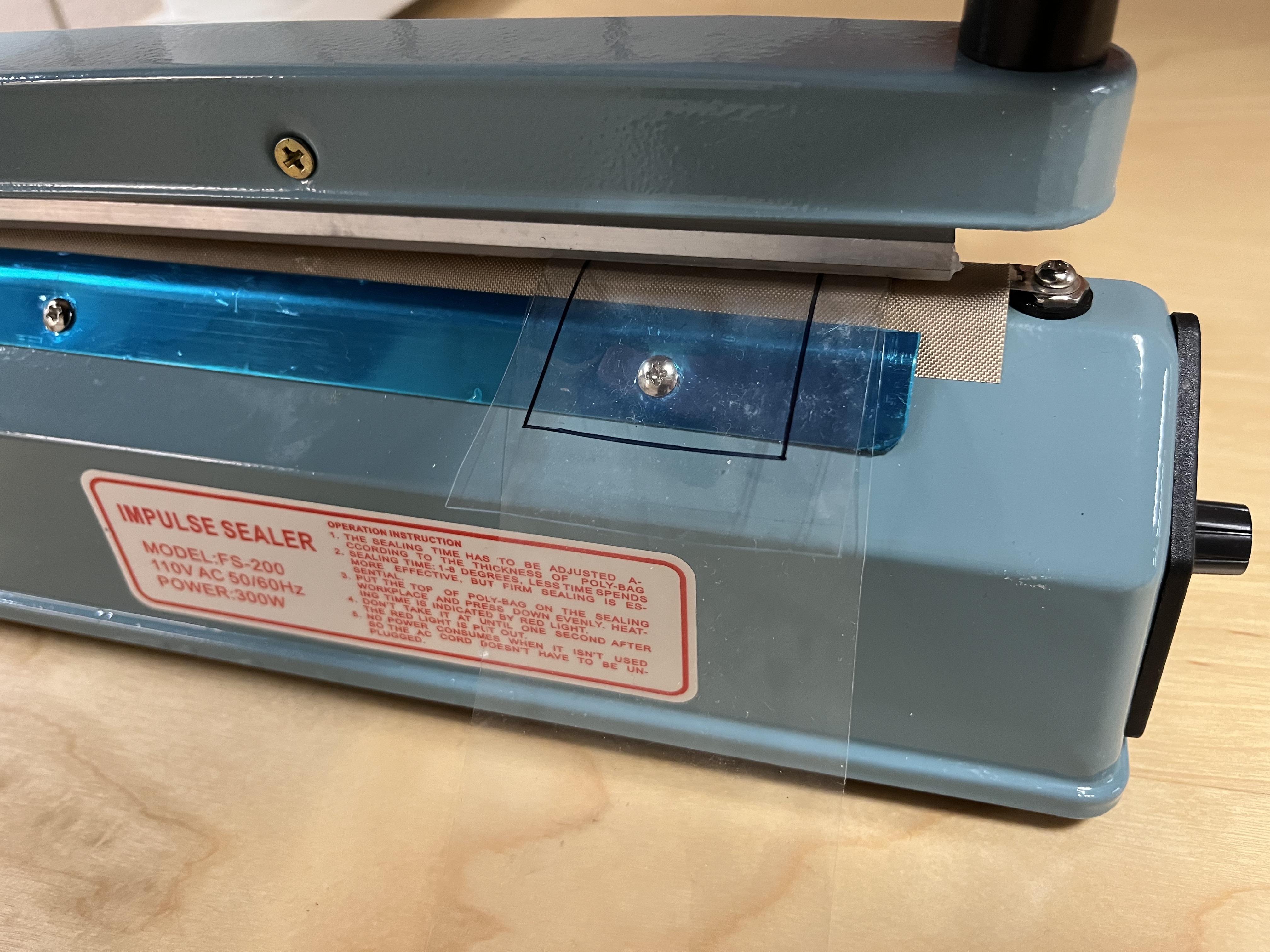}
    \caption{}
    \label{fig:bead_bag_sealed_edges}
\end{subfigure}
\begin{subfigure}{0.3\columnwidth}
    \centering
    \includegraphics[height=18mm]{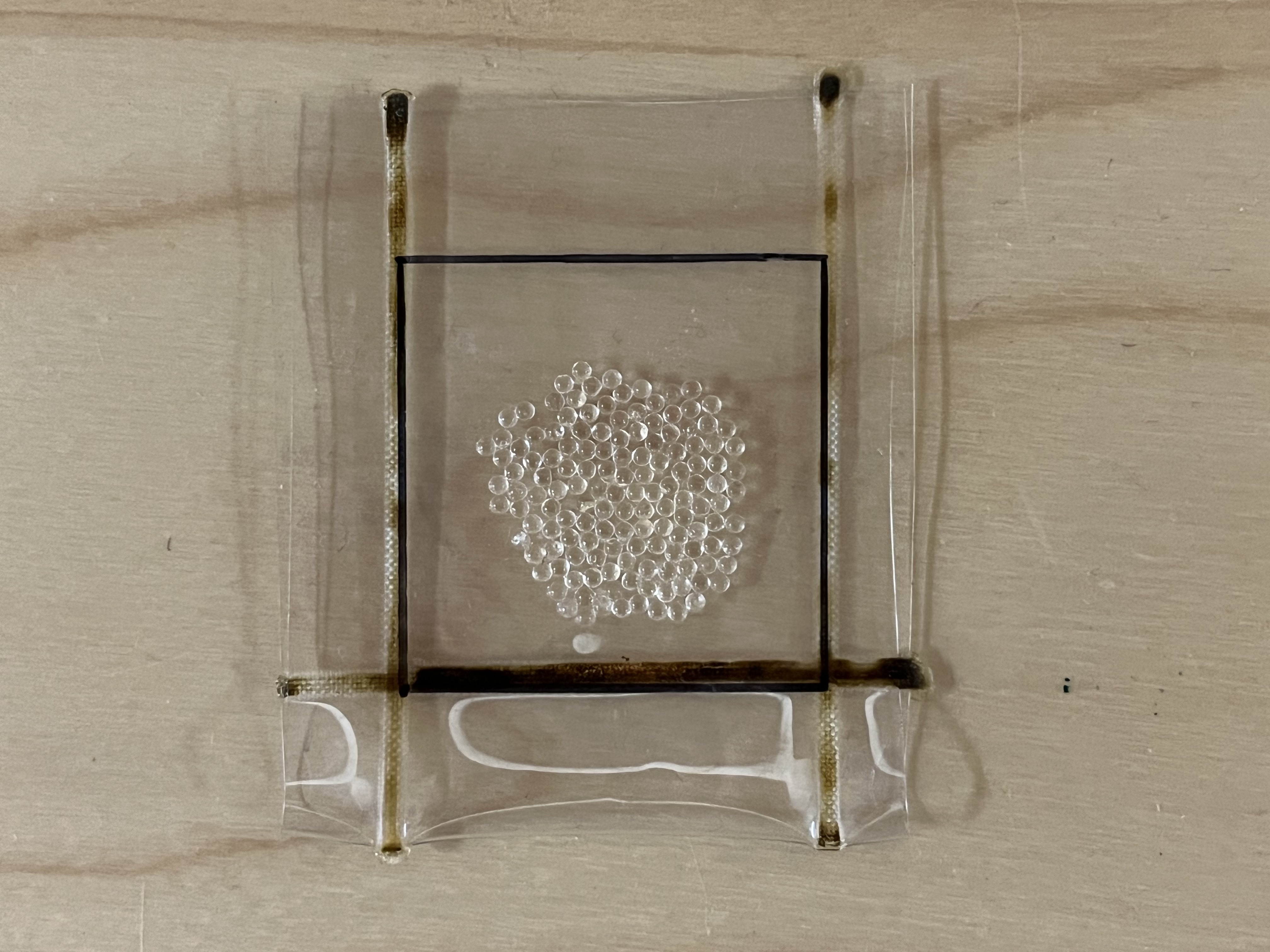}
    \caption{}
    \label{fig:bead_bag_inserted_beads}
\end{subfigure}
\hfill
\begin{subfigure}{0.3\columnwidth}
    \centering
    \includegraphics[height=18mm]{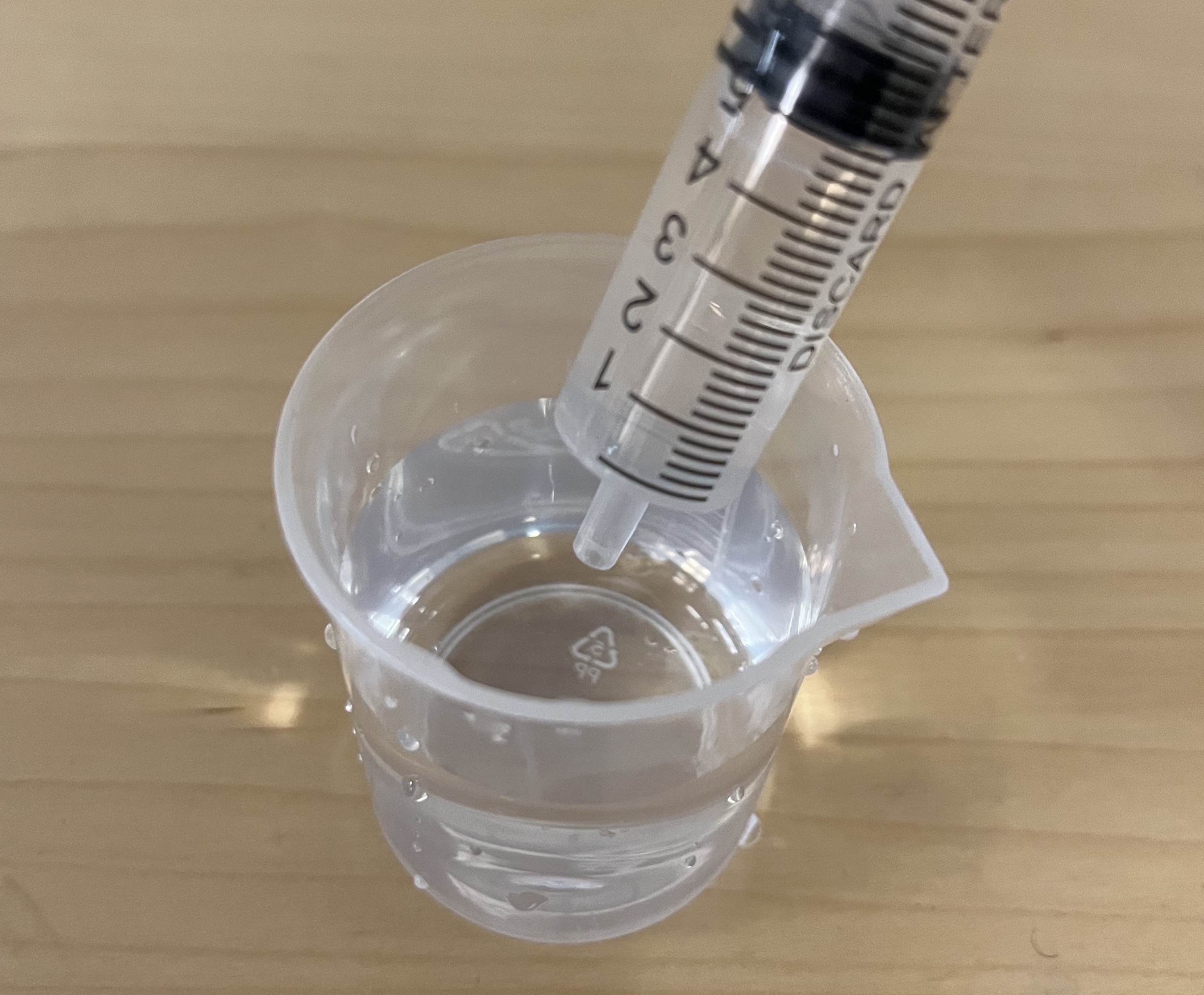}
    \caption{}
    \label{fig:bead_bag_water}
\end{subfigure}
\begin{subfigure}{0.3\columnwidth}
    \centering
    \includegraphics[height=18mm]{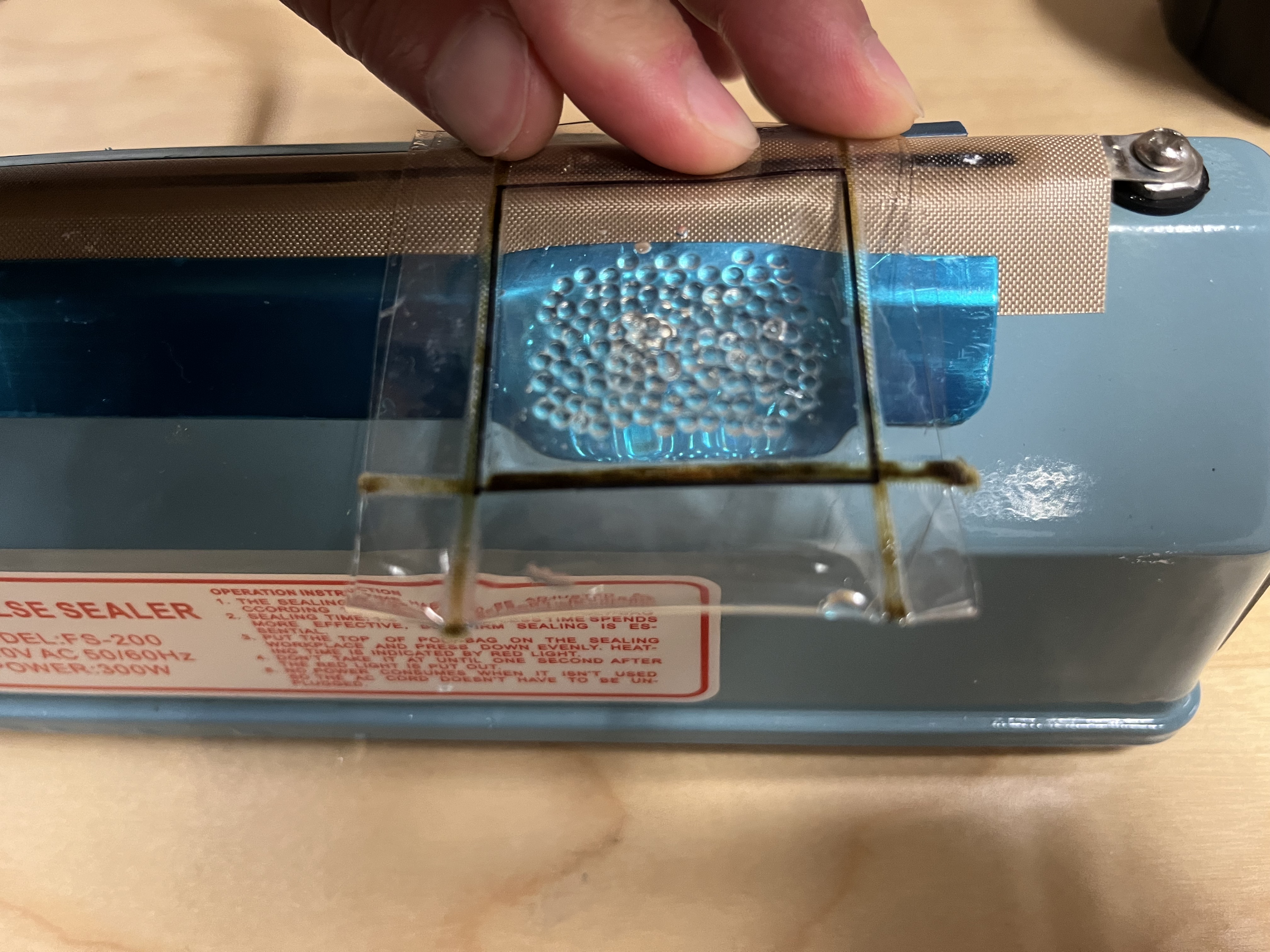}
    \caption{}
    \label{fig:bead_bag_sealed_final}
\end{subfigure}
\begin{subfigure}{0.3\columnwidth}
    \centering
    \includegraphics[height=18mm]{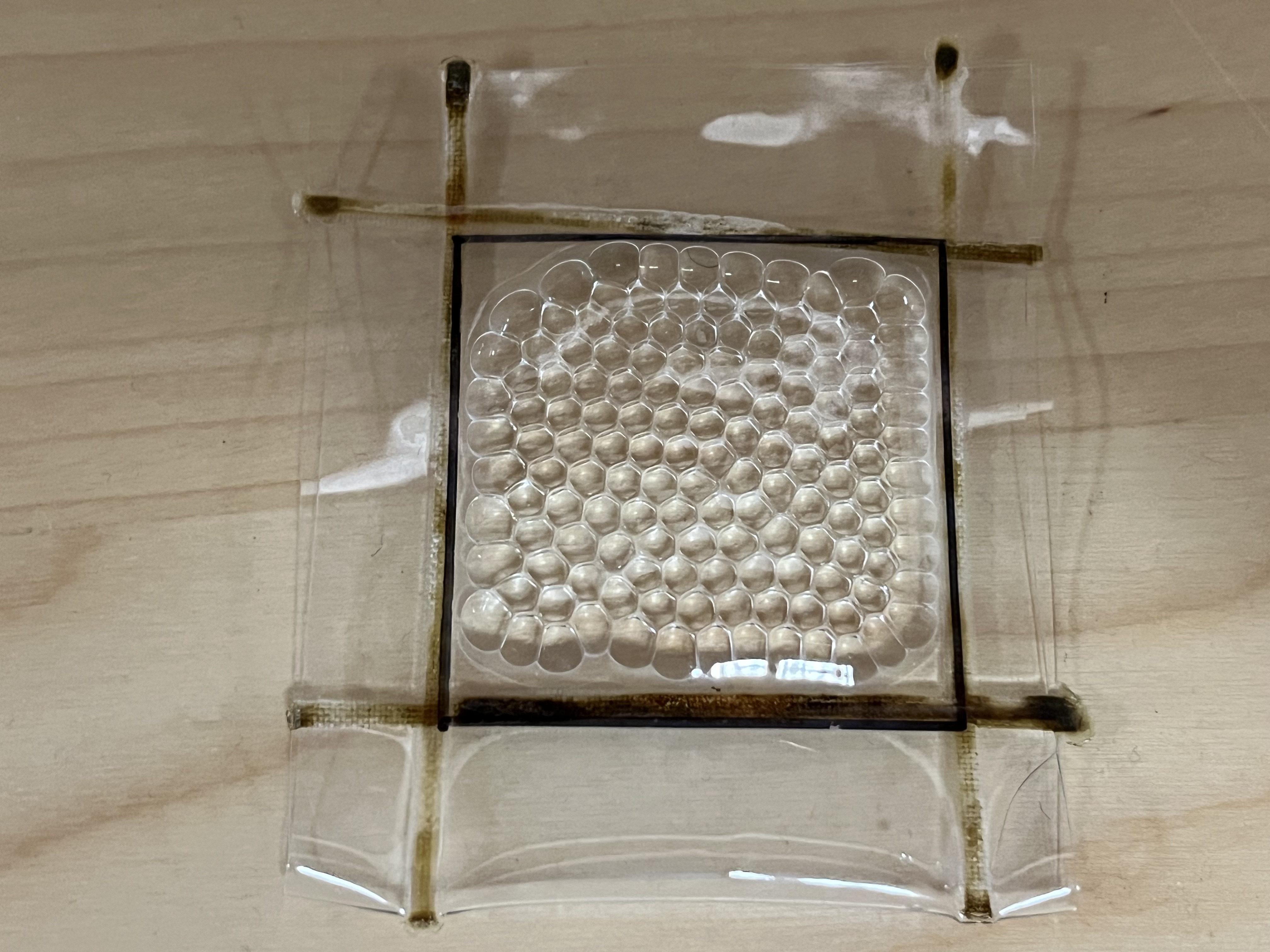}
    \caption{}
    \label{fig:bead_bag_wait_20m}
\end{subfigure}
\caption{
\textit{Bead Bag} fabrication process \textbf{(a)} 40 mm x 40 mm outline
for drawn for edges. \textbf{(b)} 3 out of the 4 edges are heat sealed.
\textbf{(c)} \textasciitilde 100 PAM beads are inserted into the bag.
\textbf{(d)} \textasciitilde 5 ml of water is injected and \textbf{(e)}
heat sealed. \textbf{(f)} Beads fully absorb water after \textasciitilde
20 minutes.
}
\label{fig:bead_bag_fabrication}
\end{figure}

To fabricate bead bags, a 40 mm x 40 mm PVC pouch was first created using two layers of PVC sheet, sealed on three edges using an impulse sealer. Once the initial envelope was formed, PAM beads were placed into the bag and 5 mL of water was injected via syringe. The final side of the pouch was then closed with the impulse sealer, and the bag was left to allow for the beads to fully saturate.  During the sealing process, a small amount of air is allowed in with the water and beads to make individual beads more visually distinctive. A detailed breakdown of this process can be seen in Fig. \ref{fig:bead_bag_fabrication}. The raw material cost to fabricate a single bead bag following this process amounts to just under \$0.09 per bag.

\subsection{Camera and Housing}

To observe the deformation of the hydrogel beads, BeadSight uses a 1080 x 1920 pixel USB camera, running at 30 Hz to capture images of the bead bag. A 180\textdegree{}  fish-eye camera lens is used, allowing the camera to be closer to the bead sack for a slimmer design, and an unwarping operation is applied to the camera stream when recording. Additionally, although the camera records at 1080x1920 pixels, we downsample to a 256x256 image before processing to decrease compute requirements. Four LED lights are mounted at the corners of the camera to illuminate the bead bag. Finally, a mounting bracket was included to affix the sensor assembly to a Franka Emika Panda robotic arm, although this mount could be easily changed to attach the sensor to alternative robotic manipulators. 

\section{Data Collection Process}\label{sec:data collection}

\subsection{Presure-map reconstruction}
When contact forces are applied to the sensor's surface, the beads are deformed and displaced. By measuring these changes, we can reconstruct the forces applied to the surface of the sensor. However, the dynamics of the beads are hard to model from first principles, as the bead behavior is a combination of solid mechanics (the deformation of individual beads and the bead sack) and fluid mechanics (the motion of beads suspended in the fluid inside the sack), with friction and adhesive forces playing a large role. As such, instead of modeling the system's dynamics, we trained a U-Net to inverse the system dynamics, learning to predict the pressure applied to the surface directly from pixel-level observations of the beads. Because applied forces cause relative motion of the beads (they have no fixed location), in order to calculate the pressure map our network must know about the relative motion of the beads. To provide this information, we pass the prior $n$ captured frames $O_{(t-n, t]}$ (we chose $n=15$) into the model. Based on these frames, the model calculates the current pressure map $P_{t}$. Our model was trained to minimize the mean squared error between the calculated pressure map and the ground truth pressure map.
$$min(||P_{t} - f_{\theta}(O_{(t-n, t]})||_2)$$

\begin{figure}[htbp]
\centering
\includegraphics[width=\columnwidth]{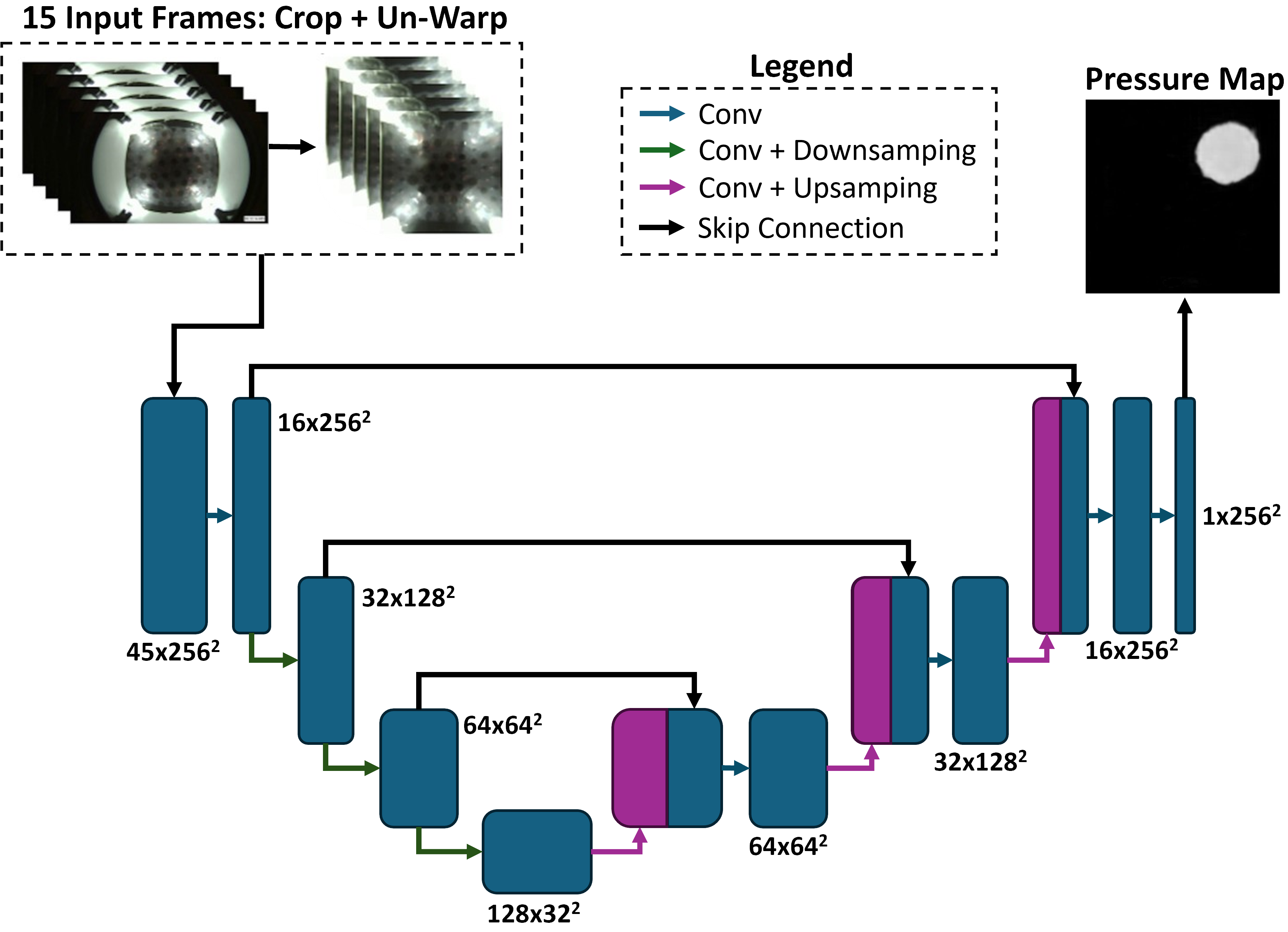}
\caption{Architecture of the UNet model. The 15 prior frames are used to calculate the current pressure map.}
\label{fig:flowchart}
\end{figure}

Our UNet's architecture consists of an encoder path with four convolutional blocks interspersed with three downsampling blocks and a decoder path with four convolutional blocks interspersed with 3 upsampling blocks, with skip connections running from the inputs of the downsampling block to the outputs of the upsampling blocks. Each convolutional block is composed of two 3x3 convolutional layers with instance normalization and ELU activation. The downsampling block uses a single 3x3 convolution (with a stride of two to halve the feature map size) and a LeakyRELU activation. The upsampling block uses nearest-neighbor interpolation to double the size of the feature map and then passes it through a single 3x3 convolutional layer. This layer also halves the number of channels, so that the skip connections can be formed by concatenating the maps from the encoder path to those from the decoder path along the channel dimension. To handle the multi-frame input, the network flattens the $n$ input frames along the channel dimension, viewing the input as a $3nx256x256$ array. The UNet outputs a single-channel pressure map of the same dimension as the input ($1x256x256$). A diagram of our U-Net can be seen in Fig. \ref{fig:flowchart}.

\subsection{Data Collection}
\begin{figure}[htbp]
\centering

\begin{subfigure}{0.46\columnwidth}
\includegraphics[width=\linewidth]{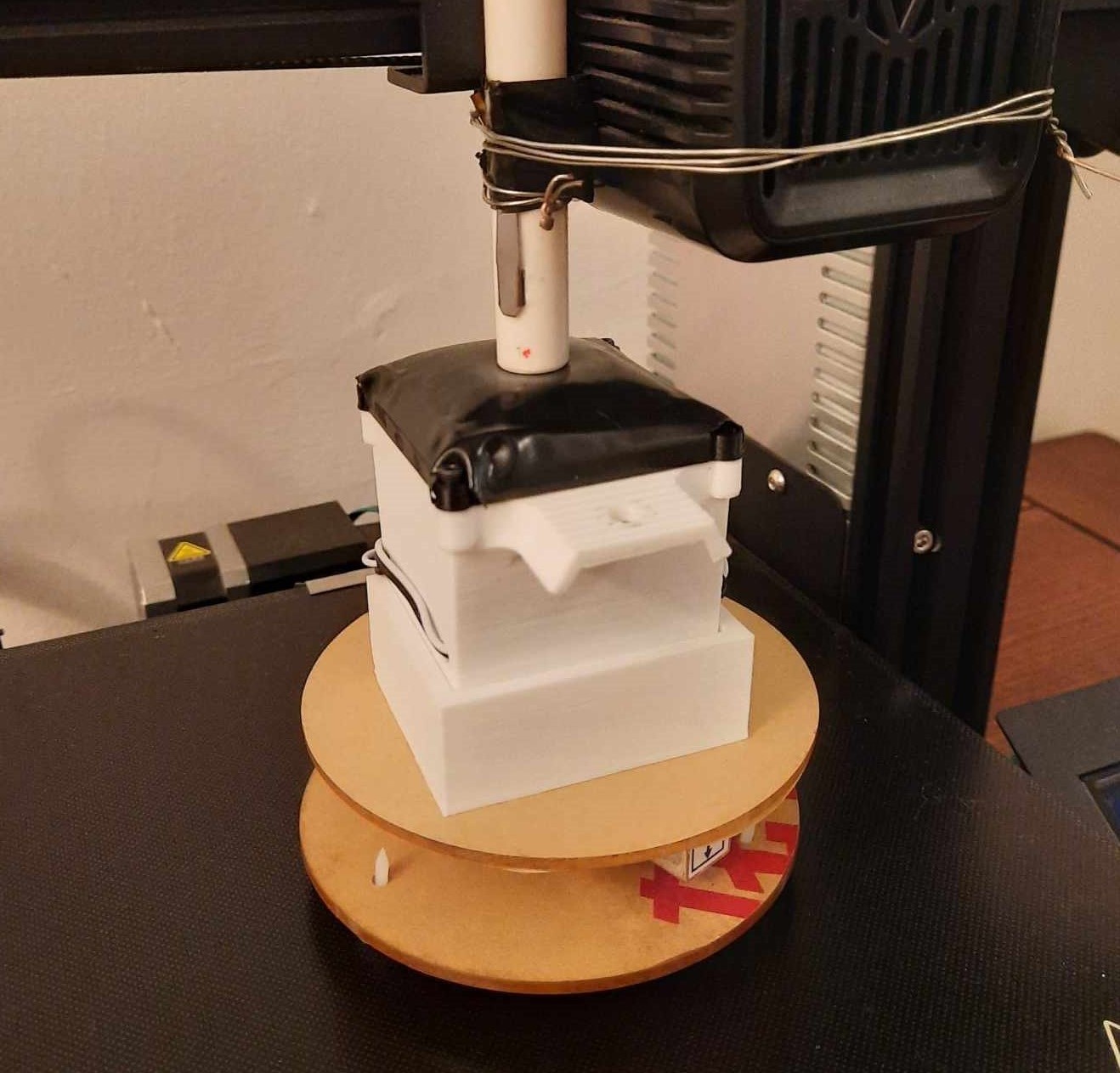}
\caption{Experimental setup}
\label{Overview1}
\end{subfigure}
\hfill
\begin{subfigure}{0.5\columnwidth}
\includegraphics[width=\linewidth]{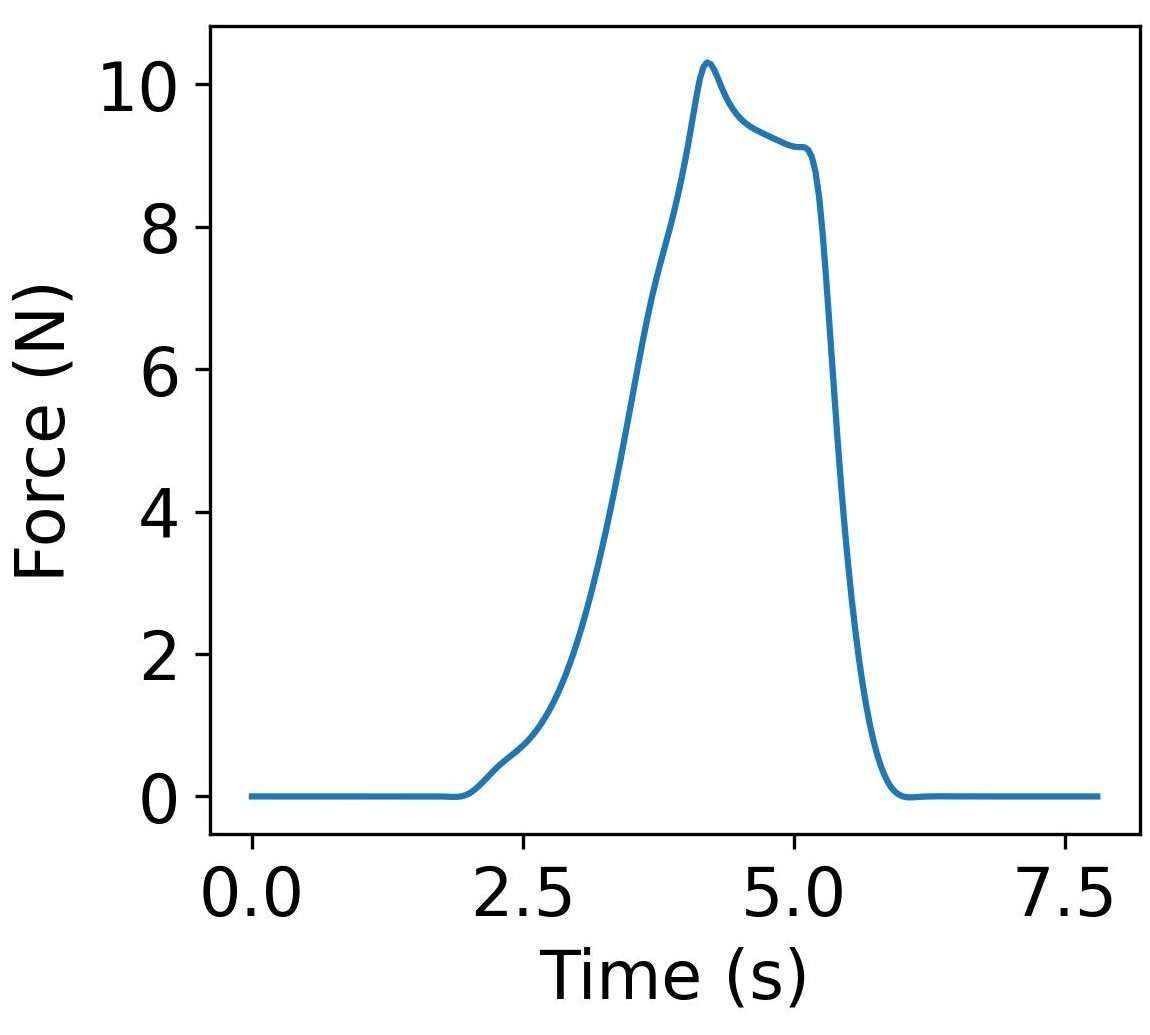}
\caption{Force over time}
\label{Overview3}
\end{subfigure}

\caption{Visualization of experimental set-up. A 3D printer is used to create cylindrical presses at random locations and random depths. The force applied is recorded by a scale below the sensor. (Left) shows a graph of total force over time for a single press. }
\label{fig:experiments}
\end{figure}

To collect training data for our model, we applied a series of controlled contacts to the sensor's surface while recording the bead displacement, contact location, and total force. Using a cylindrical "finger" attached to the head of a 3D printer, we were able to make precise contacts, recording the exact location and depth of each press. We measured the force applied by the finger over time using a digital scale. By combining the location of the finger, the contact area, and the total force over time, we were able to recreate pressure maps for each frame captured by the camera. A diagram of our experimental setup can be found in Fig. \ref{fig:experiments}. In total, we collected 500 presses (at randomized locations on the sensor), each averaging 192 frames (6.4 seconds). We applied an 70/10/20 train/validation/test split on these press episodes. To increase our data, we applied random rotations (in 90\textdegree{} increments) and flips to the training data, resulting in an octupling of our data. Using this data, we trained our U-Net model using an MSE loss and the ADAM optimizer. An outline of our training method can be seen in Algorithm 1.

\begin{algorithm}
\caption{\label{alg:logic} U-Net Training}
\label{figurelabel}
\begin{algorithmic}
\State \textbf{Input:}$\,\{O$\},\,$\{P\}$ \Comment{Visual Observations, Pressure Maps}
\State \textbf{Observation Horizon:} $H$
\State \textbf{Initialize:} $U_\theta$ \Comment{Initialize U-Net Model}

\While{Training}
    \State Sample recording $n$ from observations.
    \State Sample time $t$ from recording $n$, $H-1 \leq t < len(O_n)$
    \State $obs = O_{n, (t-H, t]}$
    \State Sample $rot$ from [0\textdegree{}, 90\textdegree{}, 180\textdegree{}, 270\textdegree{}]
    \State Sample $flip$ from [True, False]
    \State Apply a $rot$ rotation to all images in $obs$
    \State If $flip$, apply a vertical flip to all images in $obs$
    \State $Pred_{n, t}$ = $U_\theta(obs)$ \Comment{Predict Pressure Map}
    \State $loss$ = $||Pred_{n, t} - P_{n, t}||_2$ \Comment{L2 Norm}
    \State Update $U_\theta$ using $loss$ with ADAM optimizer
\EndWhile

\end{algorithmic}
\end{algorithm}

\section{Results and Discussions}\label{sec:Results}

\subsection{Pressure map Reconstruction}

\begin{figure}[htbp]
\centering
\includegraphics[width=\columnwidth]{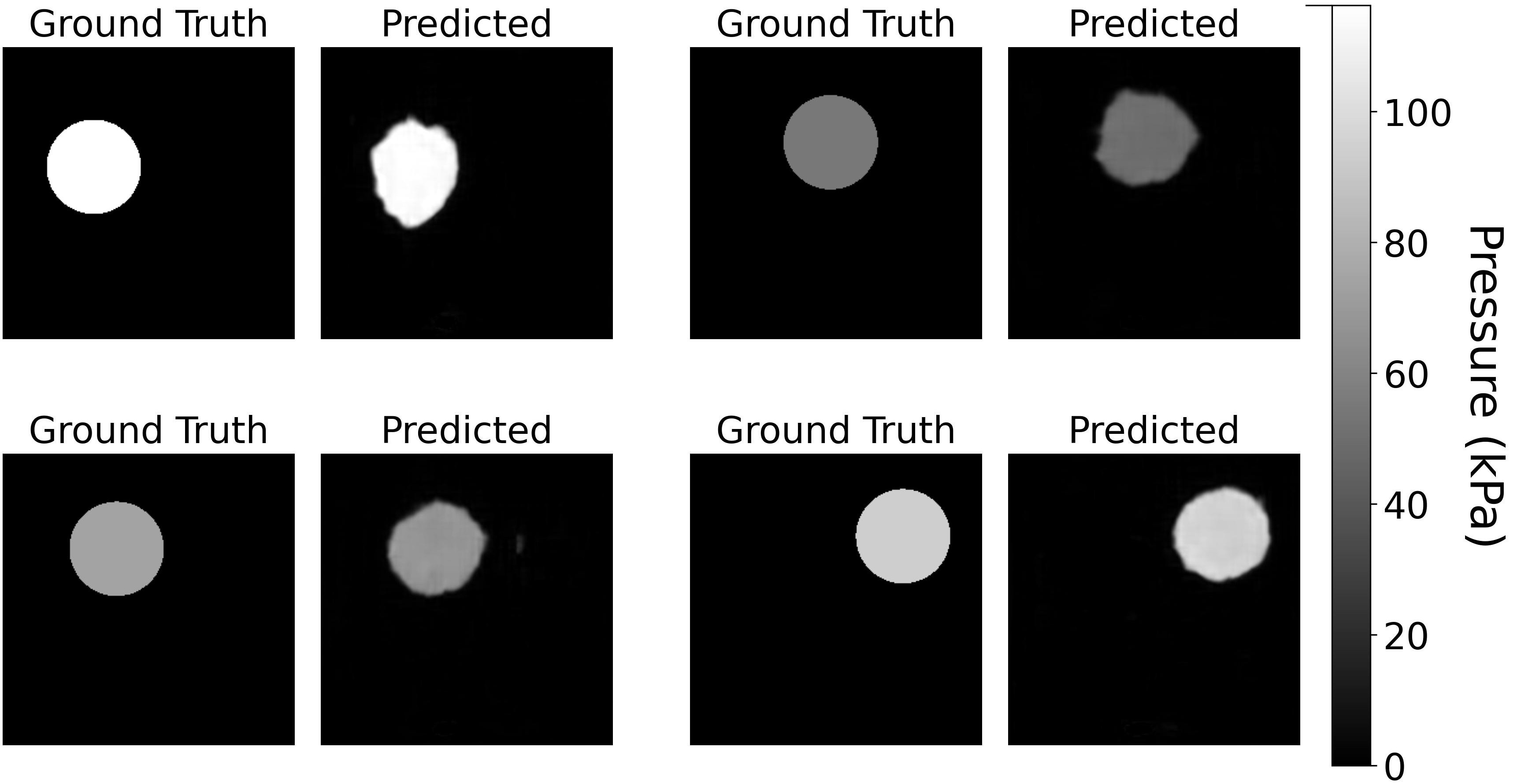}
\caption{Pressure map reconstruction prediction results for four example test cases.}
\label{fig:results}
\end{figure}

To determine the effectiveness of our pressure map reconstruction, we evaluated our trained model on our testing datasets, getting a mean absolute error (MAE) of 0.79 kPa. An example of some of these reconstructed pressure maps can be seen in Fig. \ref{fig:results}. In addition to the overall pressure map reconstruction error, we also examined the accuracy of the sensor in areas directly applicable to robotics: total force measurement accuracy, center of pressure accuracy, and intersection over union (IOU) of the contact areas. For total force measurement, we compared the total force the sensor reported (integral of the pressure map) to the recorded total force for each press and calculated the average percent MAE for all recorded pressure maps where a non-zero force was applied to the sensor. Overall, BeadSight had an 11.9\% force reconstruction error. For both the center of pressure measurements and the pressure map IOU comparisons, the finger had to be fully in contact with the sensor. To ensure this constraint was satisfied, we excluded timesteps where the total force was less than 2N of contact force (approximately 10\% of the maximum force applied during a press). Looking at this data, we found that on average the reconstructed pressure map's center of pressure was 1.18 mm from the ground truth center of pressure, meaning that BeadSight can localize contact forces with millimeter-level accuracy. Finally, we examined the IOU of the contact area, using Otzu Threshodling \cite{otsu1975threshold} to convert the pressure maps to binary contact maps. BeadSight got an average IOU of 81\%, meaning the sensor can accurately determine the contact area, in addition to the center of pressure. 

\begin{figure}[htbp]
\centering
\includegraphics[width=0.8\columnwidth]{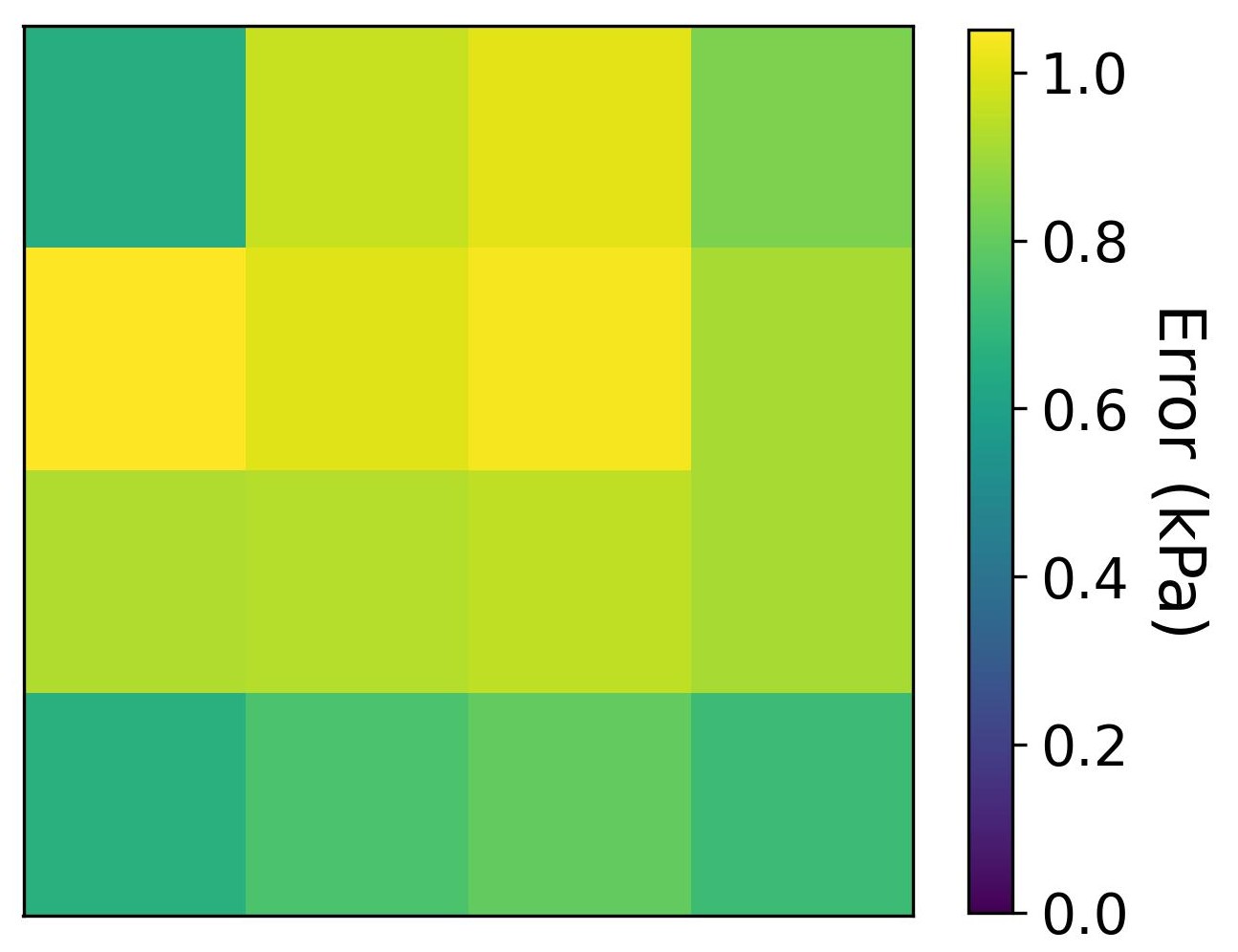}
\caption{Average absolute error for presses on different sections of the sensor.}
\label{fig:error_vs_results}
\end{figure}

To test our sensor for dead zones, we also evaluated the MAE for presses in 16 different segments of the sensor. The loss heat map from this experiment can be seen in Fig. \ref{fig:error_vs_results}. From this, we see that the sensor does perform slightly worse in the center, with decreased error on the corners and edges. We think this is likely due to the beads being more constrained along the edges of the bag, making their movements less stochastic. However, in the worse segment, the average error is only 1.05 kPa (33\% higher than the overall average error), suggesting BeadSight is effective at detecting forces across the entirety of its visible surface.

\subsection{Robot Validation}
\begin{figure}[htbp]
    \centering
    \includegraphics[width=0.49\columnwidth]{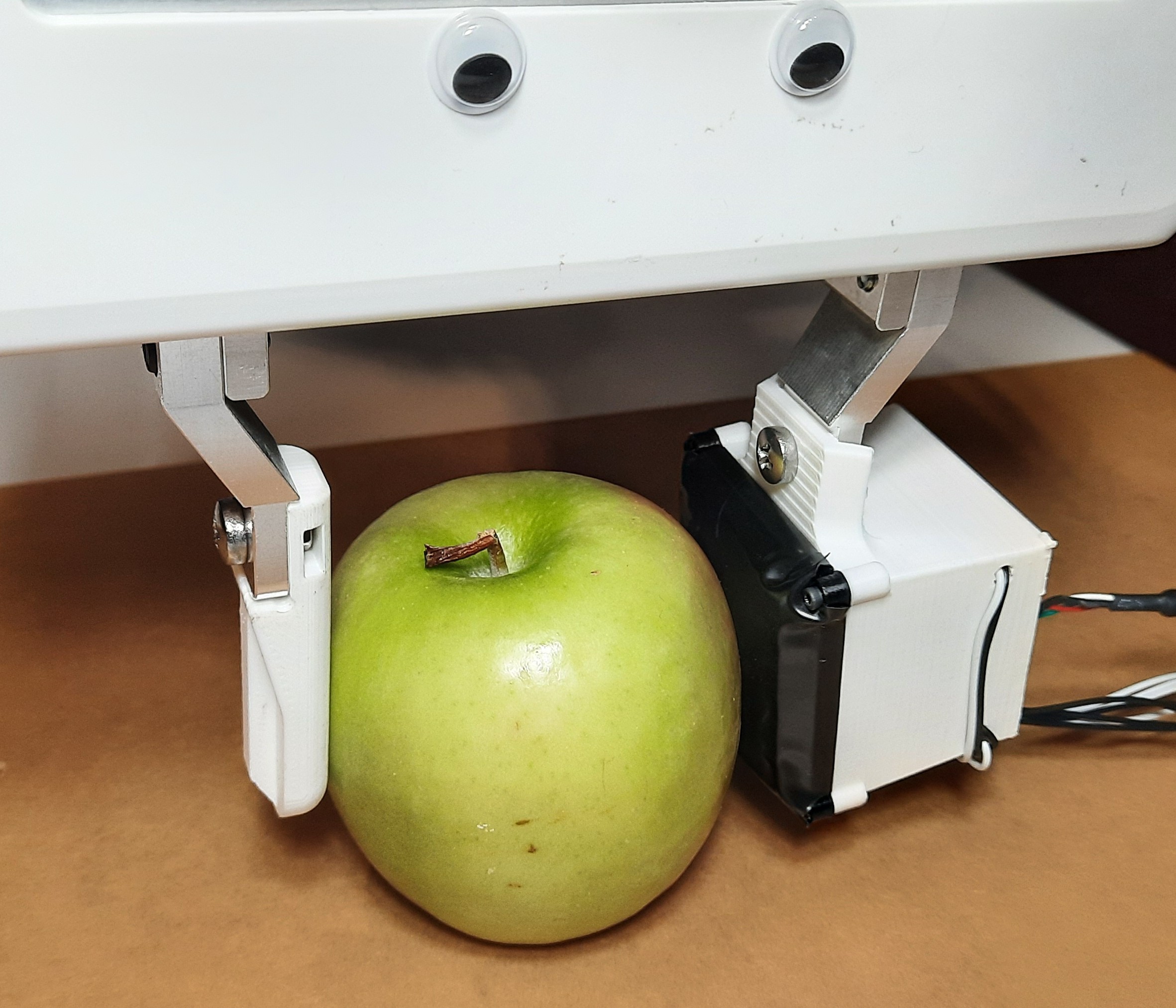}
    \includegraphics[width=0.49\columnwidth]{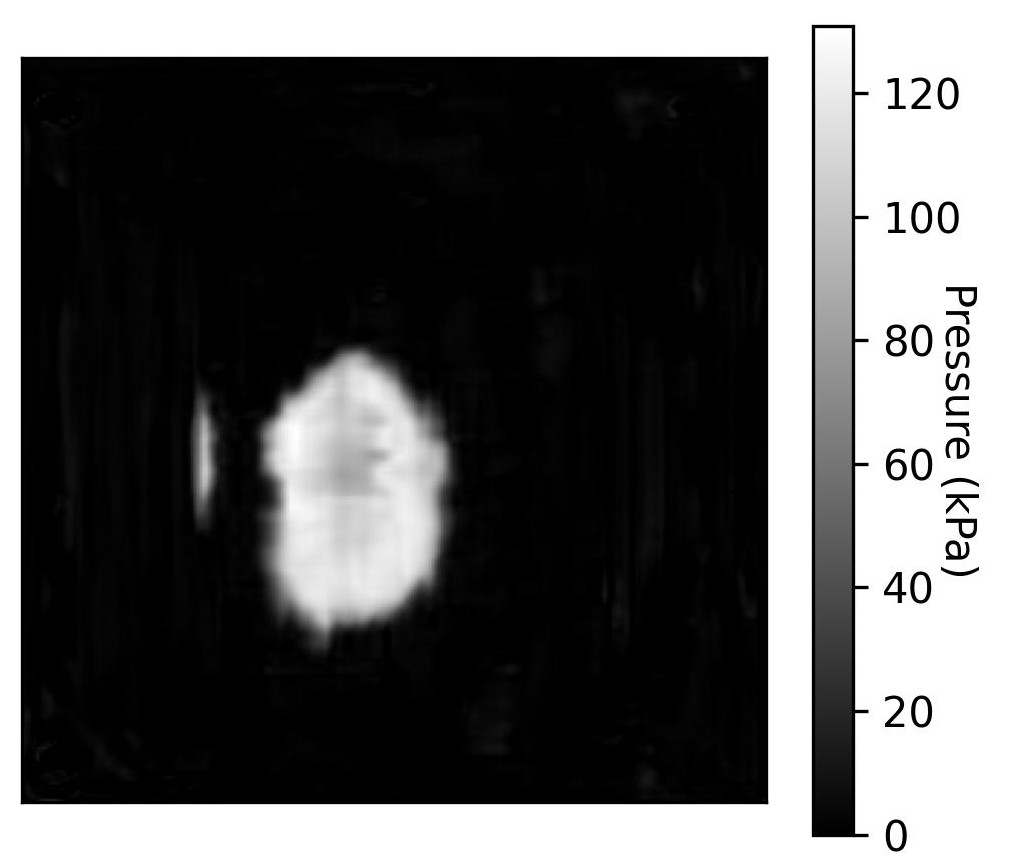}
    \caption{Hardware validation. BeadSight was mounted to a robot end-effector and used to grasp an apple. (Left) shows an example grasping action and (Right) shows a predicted pressure map from the BeadSight sensor.}
    \label{fig:robot_test}
\end{figure}

Finally, to verify that our sensor works in situ, we ran a simple verification study using a Franka Emika Panda Robot, gripping an apple. In this experiment, the Franka Emika grasped an apple using an end-effector with the BeadSight sensor mounted to it. Using the observations from the BeadSight camera, we recreated the contact pressure map, and verified that it was the correct shape and location and that the total magnitude aligned with the force applied by the Franka gripper. An image of the robot grasping the apple, along with the reconstructed pressure map, can be seen in Fig. \ref{fig:robot_test}.

Examining the pressure maps generated from the apple experiment, we found that their shapes mostly aligned with expectations, being circular and centered upon the contact point with the apple. Additionally, we found that the total force measured (integral of the pressure map) reached a maximum of 13N, approximately the gripper force of 15N.

\section{CONCLUSIONS}\label{sec:Conclusion}
In this paper, we have introduced BeadSight, a novel tactile sensor design that focuses on durability and cost-effectiveness. Despite costing less than \$60 to produce the sensor, and only cents to produce replacement sensor pads, BeadSight is an effective tactile sensor suited for robotic applications. Our results show that BeadSight is capable of highly accurate tactile readings. By training a U-Net to recreate the forces exerted on the bead sight based on pixel-only observations, BeadSight was able to predict the pressure map exerted on its surface within an error of 0.79 KPa. Although BeadSight may not be as sensitive as other visuotactile sensors, its decent performance, superior durability, and cost-effectiveness make it a valuable addition to the family of robotic sensors, especially for robots engaged in repetitive, high contact-force tasks. 

Moving forward, we hope to examine integrating BeadSight's neural-network-based perception system directly into deep learning control for robotics. In such a situation, the control neural network could take BeadSight's U-Net output as a direct input, allowing the U-Net to be fine-tuned during the training of the robotic agent. Additionally, this would help overcome the main limitation of BeadSight, which is that the U-Net's performance may suffer if the bead sight sensor is used in an application significantly different from the training application.

\section{Acknowledgments}
The authors would like to express their gratitude to Selam Gano for her assistance in constructing the BeadSight housing depicted in Fig. \ref{fig:overview} and Fig. \ref{fig:experiments} 








\newpage

\bibliographystyle{IEEEtran}
\bibliography{references}

\end{document}